\documentclass[journal]{IEEEtran}
\usepackage{cite}
\ifCLASSINFOpdf
  \usepackage[pdftex]{graphicx}
\else
  \usepackage[dvips]{graphicx}
\fi

\usepackage{amsmath}
\usepackage{algorithmic}
\usepackage{algorithm}
\usepackage{array}
\usepackage{amsfonts}
\usepackage{amssymb}
\usepackage{bm}
\usepackage{color}
\usepackage{adjustbox}

\ifCLASSOPTIONcompsoc
  \usepackage[caption=false,font=normalsize,labelfont=sf,textfont=sf]{subfig}
\else
  \usepackage[caption=false,font=footnotesize]{subfig}
\fi
\usepackage{url}

\begin{document}

\title{Low Dose CT Image Denoising Using a Generative Adversarial Network with Wasserstein Distance and Perceptual Loss}

\author{Qingsong Yang, Pingkun Yan*,~\IEEEmembership{Senior Member, IEEE}, Yanbo Zhang,~\IEEEmembership{Member, IEEE}, Hengyong Yu,~\IEEEmembership{Senior Member, IEEE}, Yongyi Shi, Xuanqin Mou,~\IEEEmembership{Senior Member, IEEE}, Mannudeep K. Kalra, Yi Zhang~\IEEEmembership{Member, IEEE}, Ling Sun, and Ge Wang,~\IEEEmembership{Fellow, IEEE}
\thanks{Asterisk indicates corresponding author.}
\thanks{This work was supported in part by the National Natural Science Foundation of China under Grant 61671312, and in part by the National Institute of Biomedical Imaging and Bioengineering/National Institutes of Health under Grant R01  EB016977 and Grant U01 EB017140.}
\thanks{The authors would also like to thank NVIDIA Corporation for the donation of the Titan Xp GPU used for this research.}
\thanks{Q. Yang, P. Yan* and G. Wang are with Department of Biomedical Engineering, Rensselaer Polytechnic Institute, Troy, NY, 12180 (e-mail: yangq4@rpi.edu, yanp2@rpi.edu, wangg6@rpi.edu)}
\thanks{Y. Zhang and H. Yu are with the Department of Electrical and Computer Engineering, University of Massachusetts Lowell, Lowell, MA 01854 (email:yanbo\_zhang@uml.edu, hengyong-yu@ieee.org)}
\thanks{Y. Shi and X. Mou are with the Institute of Image Processing and Pattern Recognition,
Xian Jiaotong University, Xian, Shaanxi 710049, China (email: xqmou@mail.xjtu.edu.cn)}
\thanks{M. K. Kalra is with Department of Radiology, Massachusetts General Hospital, Harvard Medical School, Boston, MA, USA(email: mkalra@mgh.harvard.edu) }
\thanks{Y. Zhang is with the College of Computer Science, Sichuan Universiyt, Chengdu 610065, China (e-mail: yzhang@scu.edu.cn)}
\thanks{L. Sun is with Huaxi MR Research Center (HMRRC), Department of Radiology, West China Hospital, Sichuan University, Chengdu 610041, China (e-mail: 251834489@qq.com)}
}

\maketitle

\begin{abstract}
The continuous development and extensive use of CT in medical practice has raised a public concern over the associated radiation dose to the patient. Reducing the radiation dose may lead to increased noise and artifacts, which can adversely affect the radiologist’s judgement and confidence. Hence, advanced image reconstruction from low-dose CT data is needed to improve the diagnostic performance, which is a challenging problem due to its ill-posed nature. Over the past years, various low-dose CT methods have produced impressive results. However, most of the algorithms developed for this application, including the recently popularized deep learning techniques, aim for minimizing the mean-squared-error (MSE) between a denoised CT image and the ground truth under generic penalties. Although the peak signal-to-noise ratio (PSNR) is improved, MSE- or weighted-MSE-based methods can compromise the visibility of important structural details after aggressive denoising. This paper introduces a new CT image denoising method based on the generative adversarial network (GAN) with Wasserstein distance and perceptual similarity. The Wasserstein distance is a key concept of the optimal transport theory, and promises to improve the performance of GAN. The perceptual loss suppresses noise by comparing the perceptual features of a denoised output against those of the ground truth in an established feature space, while the GAN focuses more on migrating the data noise distribution from strong to weak statistically. Therefore, our proposed method transfers our knowledge of visual perception to the image denoising task and is capable of not only reducing the image noise level but also trying to keep the critical information at the same time. Promising results have been obtained in our experiments with clinical CT images.
\end{abstract}


 \ifCLASSOPTIONpeerreview
 \begin{center} \bfseries EDICS Category: 3-BBND \end{center}
 \fi
%
\IEEEpeerreviewmaketitle

\section{Introduction}
\IEEEPARstart{X}{-ray} computed tomography (CT) is one of the most important imaging modalities in modern hospitals and clinics. However, there is a potential radiation risk to the patient, since x-rays could cause genetic damage and induce cancer in a probability related to the radiation dose~\cite{brenner2007computed,de2004risk}. Lowering the radiation dose increases the noise and artifacts in reconstructed images, which can compromise diagnostic information. Hence, extensive efforts have been made to design better image reconstruction or image processing methods for low-dose CT (LDCT). These methods generally fall into three categories: (a) sinogram filtration before reconstruction~\cite{wang2005sinogram,wang2006penalized,manduca2009projection}, (b) iterative reconstruction~\cite{beister2012iterative,hara2009iterative}, and (c) image post-processing after reconstruction~\cite{ma2011low,chen2013improving,feruglio2010block}.

Over the past decade, researchers were dedicated to developing new iterative algorithms (IR) for LDCT image reconstruction. Generally, those algorithms optimize an objective function that incorporates an accurate system model~\cite{de2004distance,lewitt1990multidimensional}, a statistical noise model~\cite{whiting2006properties,elbakri2002statistical,ramani2012splitting} and prior information in the image domain. Popular image priors include total variation (TV) and its variants~\cite{sidky2008image,liu2012adaptive,tian2011low}, as well as dictionary learning~\cite{xu2012low,zhang2017tensor}. These iterative reconstruction algorithms greatly improved image quality but they may still lose some details and suffer from remaining artifacts. Also, they require a high computational cost, which is a bottleneck in practical applications.

On the other hand, sinogram pre-filtration and image post-processing are computationally efficient compared to iterative reconstruction. Noise characteristic was well modeled in the sinogram domain for sinogram-domain filtration. However, sinogram data of commercial scanners are not readily available to users, and these methods may suffer from resolution loss and edge blurring. Sinogram data need to be carefully processed, otherwise artifacts may be induced in the reconstructed images.

Differently from sinogram denoising, image post-processing directly operates on an image. Many efforts were made\textbf{•} in the image domain to reduce LDCT noise and suppress artifacts. For example, the non-local means (NLM) method was adapted for CT image denoising~\cite{ma2011low}.
Inspired by compressed sensing methods, an adapted K-SVD method was proposed~\cite{chen2013improving} to reduce artifacts in CT images. The block-matching 3D (BM3D) algorithm was used for image restoration in several CT imaging tasks~\cite{feruglio2010block,kang2013image}. With such image post-processing, image quality improvement was clear but over-smoothing and/or residual errors were often observed in the processed images. These issues are difficult to address, given the non-uniform distribution of CT image noise.

The recent explosive development of deep neural networks suggests new thinking and huge potential for the medical imaging field~\cite{wang2017machine,wang2016perspective}. As an example, the LDCT denoising problem can be solved using deep learning techniques. Specifically, the convolutional neural network (CNN) for image super-resolution \cite{DongLHT16} was recently adapted for low-dose CT image denoising~\cite{Chen1610.00321}, with a significant performance gain. Then, more complex networks were proposed to handle the LDCT denoising problem such as the RED-CNN in~\cite{chen_zhang_kalra_lin_chen_liao_zhou_wang_2017} and the wavelet network in~\cite{kang2016deep}. The wavelet network adopted the shortcut connections introducted by the U-net~\cite{ronneberger2015u} directly and the RED-CNN [27] replaced the pooling/unpooling layers of U-net with convolution/deconvolution pairs.

Despite the impressive denoising results with these innovative network structures, they fall into a category of an end-to-end network that typically uses the mean squared error (MSE) between the network output and the ground truth as the loss function. As revealed by the recent work \cite{Johnson1603,Ledig1609}, this per-pixel MSE is often associated with over-smoothed edges and loss of details. As an algorithm tries to minimize per-pixel MSE, it overlooks subtle image textures/signatures critical for human perception. It is reasonable to assume that CT images distribute over some manifolds. From that point of view, the MSE based approach tends to take the mean of high-resolution patches using the Euclidean distance rather than the geodesic distance. Therefore, in addition to the blurring effect, artifacts are also possible such as non-uniform biases.

To tackle the above problems, here we propose to use a generative adversarial network (WGAN)~\cite{arjovsky2017wasserstein} with the Wasserstein distance as the discrepancy measure between distributions and a perceptual loss that computes the difference between images in an established feature space~\cite{Johnson1603,Ledig1609}.

The use of WGAN is to encourage that denoised CT images share the same distribution as that of normal dose CT (NDCT) images. In the GAN framework, a generative network $G$ and a discriminator network $D$ are coupled tightly and trained simultaneously. While the $G$ network is trained to produce realistic images $G(\bm{z})$ from a random vector $\bm{z}$, the $D$ network is trained to discriminate between real and generated images~\cite{goodfellow2014generative,goodfellow2016nips}. GANs have been used in many applications such as single image super-resolution~\cite{Johnson1603}, art creation~\cite{brock2016neural,zhu2016generative}, and image transformation~\cite{isola2016image}. In the field of medical imaging, Nie~\textit{et al.}~\cite{nie2016medical} proposed to use GAN to estimate CT image from its corresponding MR image. Wolterink \textit{et al.}~\cite{wolterink2017generative} are the first to apply GAN network for cardiac CT image denoising. And Yu \textit{et al.}~\cite{yu2017deep} used GAN network to handle the de-alising problem for fast CS-MRI. Promising results were achieved in these works. We will discuss and compare the results of those two networks in Section~\ref{sec:exp} since the proposed network is closely related with their works.

Despite its success in these areas, GANs still suffer from a remarkable difficulty in training~\cite{goodfellow2016nips,arjovsky2017towards}. In the original GAN \cite{goodfellow2014generative}, $D$ and $G$ are trained by solving the following minimax problem 
\begin{multline}
\min_G\max_D L_{\mathrm{GAN}}(D,G)=\mathbb{E}_{\bm{x}\sim P_r}[\log D(\bm{x})] \\ +\mathbb{E}_{\bm{z}\sim P_z}[\log \left(1- D(G(\bm{z}))\right)]
\label{eq:loss_D_GAN}
\end{multline}
where $\mathbb{E}(\cdot)$ denotes the expectation operator; $P_r$ and $P_z$ are the real data distribution and the noisy data distribution. The generator $G$ transforms a noisy sample to mimic a real sample, which defines a data distribution, denoted by $P_g$. When $D$ is trained to become an optimal discriminator for a fixed $G$, the minimization search for $G$ is equivalent to minimizing the Jensen-Shannon (JS) divergence of $P_r$ and $P_g$, which will lead to vanished gradient on the generator $G$~\cite{arjovsky2017towards} and $G$ will stop updating as the training continues. 

Consequently, Arjovsky~\textit{et al.} \cite{arjovsky2017wasserstein} proposed to use the \textit{Earth-Mover} (EM) distance or Wasserstein metric between the generated image samples and real data for GAN, which is referred to as WGAN, because the EM distance is continuous and differentiable almost everywhere under some mild assumptions while neither KL nor JS divergence is. After that, an improved WGAN with \textit{gradient penalty} was proposed~\cite{gulrajani2017improved} to accelerate the convergence. 

The rationale behind the perceptual loss is two-fold. First, when a person compares two images, the perception is not performed pixel-by-pixel. Human vision actually extracts and compares features from images \cite{Nixon2008}. Therefore, instead of using pixel-wise MSE, we employ another pre-trained deep CNN (the famous VGG~\cite{VGG}) for feature extraction and compare the denoised output against the ground truth in terms of the extracted features. Second, from a mathematical point of view, CT images are not uniformly distributed in a high-dimensional Euclidean space. They reside more likely in a low-dimensional manifold. With MSE, we are not measuring the intrinsic similarity between the images, but just their superficial differences in the brute-force Euclidean distance. By comparing images according their intrinsic structures, we should project them onto a manifold and calculate the geodesic distance instead. Therefore, the use of the perceptual loss for WGAN should facilitate producing results with not only lower noise but also sharper details. 
 
In particular, we treat the LDCT denoising problem as a transformation from LDCT to NDCT images. WGAN provides a good distance estimation between the denoised LDCT and NDCT image distributions. Meanwhile, the VGG-based perceptual loss tends to keep the image content after denoising. The rest of this paper is organized as follows. The proposed method is described in Section~\ref{sec:method}. The experiments and results are presented in Section~\ref{sec:exp}. Finally, relevant issues are discussed and a conclusion is drawn in Section~\ref{sec:conclusions}.

\section{Methods}
\label{sec:method}

\subsection{Noise Reduction Model}
Let $\bm{z} \in \mathbb{R}^{N \times N}$ denote a LDCT image and $\bm{x} \in \mathbb{R}^{N \times N}$ denote the corresponding NDCT image. The goal of the denoising process is to seek a function $G$ that maps LDCT $\bm{z}$ to NDCT $\bm{x}$:
\begin{equation}
G:\bm{z} \to \bm{x}
\end{equation}
On the other hand, we can also take $\bm{z}$ as a sample from the LDCT image distribution $P_L$ and $\bm{x}$ from the NDCT distribution or the real distribution $P_r$. The denoising function $G$ maps samples from $P_L$ into a certain distribution $P_g$. By varying the function $G$, we aim to change $P_g$ to make it close to $P_r$. In this way, we treat the denoising operator as moving one data distribution to another.

Typically, noise in x-ray photon measurements can be simply modeled as the combination of Poisson quantum noise and Gaussian electronic noise. On the contrary, in the reconstructed images, the noise model is usually complicated and non-uniformly distributed across the whole image. Thus there is no clear clue that indicates how data distributions of NDCT and LDCT images are related to each other, which makes it difficult to denoise LDCT images using traditional methods. However, this uncertainty of noise model can be ignored in deep learning denoising because a deep neural network itself can efficiently learn high-level features and a representation of data distribution from modest sized image patches through a neural network.

\subsection{WGAN}

Compared to the original GAN network, WGAN uses the Wasserstein distance instead of the JS divergence to compare data distributions. It solves the following minimax problem to obtain both $D$ and $G$ \cite{gulrajani2017improved}:
\begin{multline}
\min_{G}\max_{D}L_{\mathrm{WGAN}}(D,G) = -\mathbb{E}_{\bm{x}}[D(\bm{x})]+\mathbb{E}_{\bm{z}}[D(G(\bm{z}))] \\ +\lambda \mathbb{E}_{\hat{\bm{x}}}[(||\nabla_{\hat{\bm{x}}}D(\hat{\bm{x}})||_2-1)^2],
\label{eq: loss_DG_WGAN}
\end{multline}
where the first two terms perform a Wasserstein distance estimation; the last term is the gradient penalty term for network regularization; $\hat{\bm{x}}$ is uniformly sampled along straight lines connecting pairs of generated and real samples; and $\lambda$ is a constant weighting parameter. Compared to the original GAN, WGAN removes the $\log$ function in the losses and also drops the last sigmoid layer in the implementation of the discriminator $D$. Specifically, the networks $D$ and $G$ are trained alternatively by fixing one and updating the other.

\subsection{Perceptual Loss}
While the WGAN network encourages that the generator transforms the data distribution from high noise to a low noise version, another part of the loss function is added for the network to keep image details or information content. Typically, a mean squared error (MSE) loss function is used, which tries to minimize the pixel-wise error between a denoised patch $G(\bm{z})$ and a NDCT image patch $\bm{x}$ as~\cite{Chen1610.00321,chen_zhang_kalra_lin_chen_liao_zhou_wang_2017}
\begin{equation}
L_{\mathrm{MSE}}(G) = \mathbb{E}_{(\bm{x},\bm{z})} \left[ \frac{1}{N^2} ||G\left(\bm{z}\right)-\bm{x}||_F^2 \right],
\label{eq: mse_loss}
\end{equation}
where $||\cdot||_F$ denotes the Frobenius norm.
However, the MSE loss can potentially generate blurry images and cause the distortion or loss of details. Thus, instead of using a MSE measure, we apply a perceptual loss function defined in a feature space
\begin{equation}
L_{\mathrm{Perceptual}}(G) = \mathbb{E}_{(\bm{x},\bm{z})} \left[ \frac{1}{whd}||\phi(G(\bm{z}))-\phi(\bm{x})||_F^2 \right],
\end{equation} 
where $\phi$ is a feature extractor, and $w$, $h$, and $d$ stand for the width, height and depth of the feature space, respectively. In our implementation, we adopt the well-known pre-trained VGG-19 network \cite{VGG} as the feature extractor. Since the pre-trained VGG network takes color images as input while CT images are in grayscale, we duplicated the CT images to make RGB channels before they are fed into the VGG network. The VGG-19 network contains 16 convolutional layers followed by 3 fully-connected layers. The output of the 16th convolutional layer is the feature extracted by the VGG network and used in the perceptual loss function,
\begin{equation}
L_{\mathrm{VGG}}(G) = \mathbb{E}_{(\bm{x},\bm{z})} \left[ \frac{1}{whd}||VGG(G(\bm{z}))-VGG(\bm{x})||_F^2 \right]
\label{eq: vgg_loss}
\end{equation}
For convenience, we call the perceptual loss computed by VGG network \textit{VGG loss}. 

Combining Eqs.~(\ref{eq: loss_DG_WGAN}) and~(\ref{eq: vgg_loss}) together, we get the overall joint loss function expressed as 
\begin{equation}
\min_{G}\max_{D} L_{\mathrm{WGAN}}(D,G)+\lambda_1 L_{\mathrm{VGG}}(G)
\label{eq: loss_overall}
\end{equation}
where $\lambda_1$ is a weighting parameter to control the trade-off between the WGAN adversarial loss and the VGG perceptual loss.

\subsection{Network Structures}

\begin{figure}[t]
\centering
\includegraphics[width=.45\textwidth]{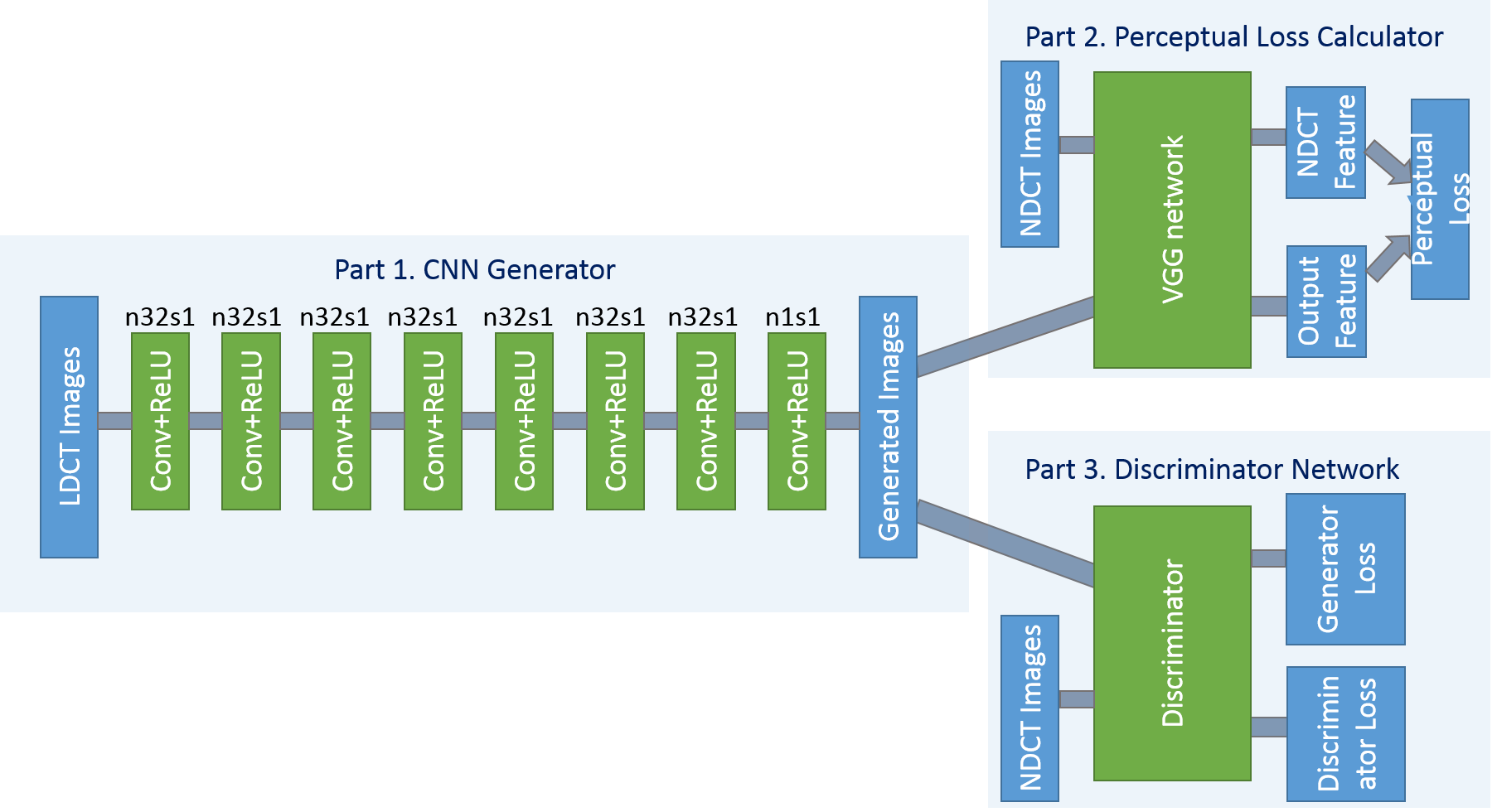}
\caption{The overall structure of the proposed WGAN-VGG network. In Part 1, $n$ stands for the number of convolutional kernels and $s$ for convolutional stride. So, $n32s1$ means the convolutional layer has 32 kernels with stride 1.}
\label{fig: overall}
\end{figure}

The overall view of the proposed network structure is shown in Fig. \ref{fig: overall}. For convenience, we name this network WGAN-VGG. It consists three parts. The first part is the generator $G$, which is a convolutional neural network (CNN) of 8 convolutional layers. Following the common practice in the deep learning community \cite{taxonomy2016}, small $3\times3$ kernels were used in each convolutional layer. Due to the stacking structure, such a network can cover a large enough receptive field efficiently. Each of the first 7 hidden layers of $G$ have 32 filters. The last layer generates only one feature map with a single $3\times3$ filter, which is also the output of $G$. We use Rectified Linear Unit (ReLU) as the activation function. 

The second part of the network is the perceptual loss calculator, which is realized by the pre-trained VGG network \cite{VGG}. A denoised output image $G(\bm{z})$ from the generator $G$ and the ground truth image $\bm{x}$ are fed into the pre-trained VGG network for feature extraction. Then, the objective loss is computed using the extracted features from a specified layer according to Eq.~(\ref{eq: vgg_loss}). The reconstruction error is then back-propagated to update the weights of $G$ only, while keeping the VGG parameters intact. 

\begin{figure}[t]
\centering
\includegraphics[width=.45\textwidth]{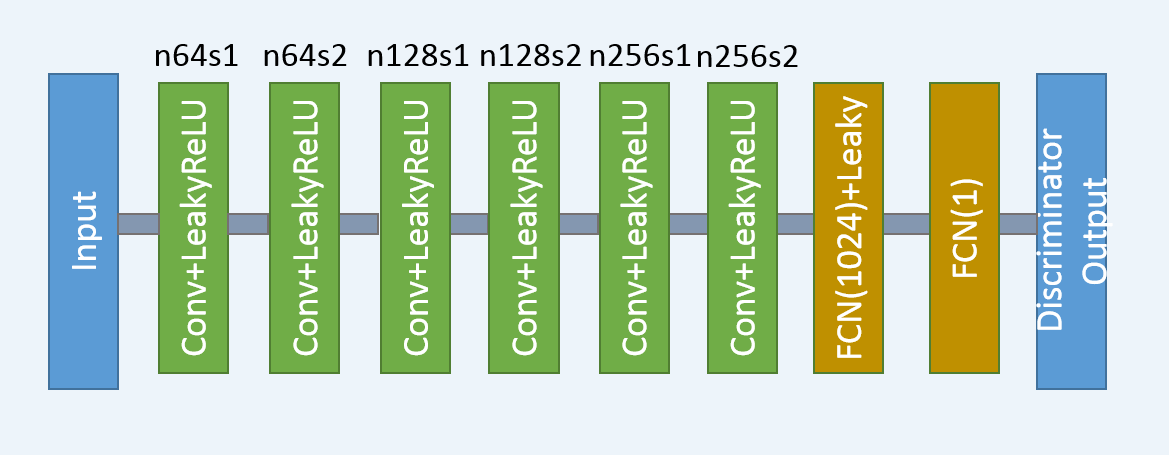}
\caption{The structure of the discriminator network. $n$ and $s$ have the same meaning as in Fig.~\ref{fig: overall}}
\label{fig: discriminator}
\end{figure}

The third part of the network is the discriminator $D$. As shown in Fig. \ref{fig: discriminator}, $D$ has 6 convolutional layers with the structure inspired by others' work \cite{VGG,Johnson1603,Ledig1609}. The first two convolutional layers have 64 filters, then followed by two convolutional layers of 128 filters, and the last two convolutional layers have 256 filters. Following the same logic as in $G$, all the convolutional layers in $D$ have a small $3\times3$ kernel size. After the six convolutional layers, there are two fully-connected layers, of which the first has 1024 outputs and the other has a single output. Following the practice in \cite{arjovsky2017wasserstein}, there is no sigmoid cross entropy layer at the end of $D$.

The network is trained using image patches and applied on entire images. The details are provided in Section~\ref{sec:exp} on experiments.

\begin{figure}
\begin{algorithmic}[1]
 \REQUIRE Set hyper-parameters, $\lambda=10, \alpha=1\times10^{-5}, \beta_1=0.5, \beta_2=0.9, \lambda_1=0.1, \lambda_2=0.1$,
 \REQUIRE Set the number of total epochs, $N_{epoch}=100$, the number of iteration for discriminator training, $N_D=4$, the batch size $m=128$, and image patch size of $80\times80$.
 \REQUIRE Initial discriminator parameters $w_0$, initial generator parameters $\theta_0$
 \REQUIRE Load VGG-19 network parameters
  \FOR {$num\_epoch = 0,...,N_{epoch}$}
  	\FOR {$t = 1,...,N_D$}
  	\STATE Sample a batch of NDCT image patches $\{\bm{x}^{(i)}\}_{i=1}^m$, latent LDCT patches $\{\bm{z}^{(i)}\}_{i=1}^m$, and random numbers $\{\epsilon^{(i)}\}_{i=1}^m \sim \mathrm{Uniform}[0,1]$
  	\FOR {$i = 1,...,m$}
  		\STATE $\hat{\bm{x}}^{(i)} \gets \epsilon^{(i)} \bm{x}^{(i)} + (1-\epsilon^{(i)})G(\bm{z}^{(i)})$
  		\STATE $L^{(i)}(D) \gets D(G(\bm{z}^{(i)})) - D(\bm{x}^{(i)}) +  \lambda(||\nabla D(\hat{\bm{x}}^{(i)})||_2-1)^2$
  	\ENDFOR
  	\ENDFOR
  	\STATE Update $D$: $w \gets \text{Adam}(\nabla_w\frac{1}{m}\sum_{i=1}^m L^{(i)}(D), w, \alpha, \beta_1, \beta_2)$
  	\STATE Sample a batch of LDCT patches $\{\bm{z}^{(i)}\}_{i=1}^m$ and corresponding NDCT patches $\{\bm{x}^{(i)}\}_{i=1}^m$, 
  	\FOR {$i = 1,...,m$}
  		\STATE $L^{(i)}(G) \gets \lambda_1 L_{\mathrm{VGG}}(\bm{z}^{(i)}, \bm{x}^{(i)}) - D(G(\bm{z}^{(i)}))$
  	\ENDFOR
  	\STATE Update $G$, $\theta \gets \text{Adam}(\nabla_{\theta}\frac{1}{m}\sum_{i=1}^m L^{(i)}(G), w, \alpha, \beta_1, \beta_2)$
  \ENDFOR
\end{algorithmic}
\caption{Optimization procedure of WGAN-VGG network.}
\label{Alg1}
\end{figure}

\subsection{Other Networks}

For comparison, we also trained four other networks.
\begin{itemize}
\item CNN-MSE with only MSE loss 

\item CNN-VGG with only VGG loss 

\item WGAN-MSE with MSE loss in the WGAN framework

\item WGAN with no other additive losses 
\item Original GAN 
\end{itemize}
All the trained networks are summarized in Table.~\ref{table:networks}.

\begin{table}[t]
\renewcommand{\arraystretch}{1.3}
\centering
\caption{Summary of all trained networks: their loss functions and trainable networks.}
\begin{tabular}{l l l}
\hline
Network & Loss \\
\hline
CNN-MSE  & $\min_G L_{\mathrm{MSE}}(G)$   \\
WGAN-MSE  & $\min_G\max_G L_{\mathrm{WGAN}}(G, D) + \lambda_2 L_{\mathrm{MSE}}(G)$\\
CNN-VGG  & $\min_G L_{\mathrm{VGG}}(G)$  \\
WGAN-VGG  & $\min_G\max_G L_{\mathrm{WGAN}}(G, D) + \lambda_1 L_{\mathrm{VGG}}(G)$ \\
WGAN  & $\min_G\max_G L_{\mathrm{WGAN}}(G, D)$ \\
GAN  & $\min_G\max_G L_{\mathrm{GAN}}(G, D)$ \\
\hline
\end{tabular}
\label{table:networks}
\end{table}

\begin{figure}[t]
\centering
\subfloat[]{\includegraphics[width=1.7in]{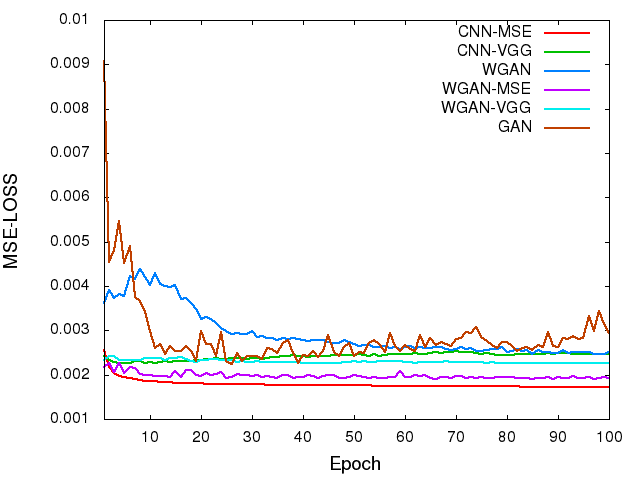}
\label{fig: mse_loss_convergence}}
\subfloat[]{\includegraphics[width=1.7in]{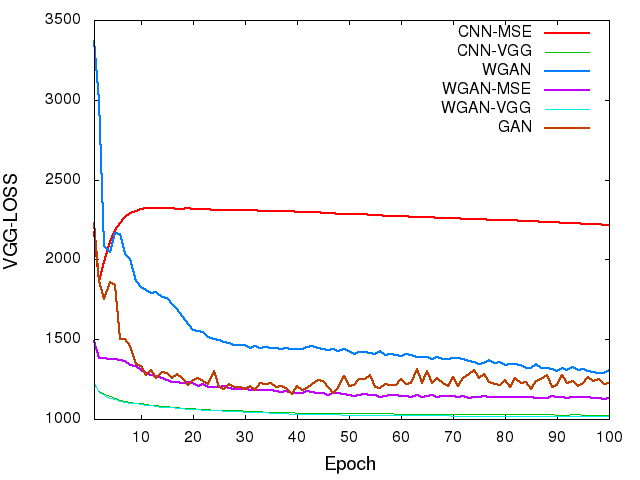}
\label{fig: vgg_loss_convergence}}

\subfloat[]{\includegraphics[width=1.7in]{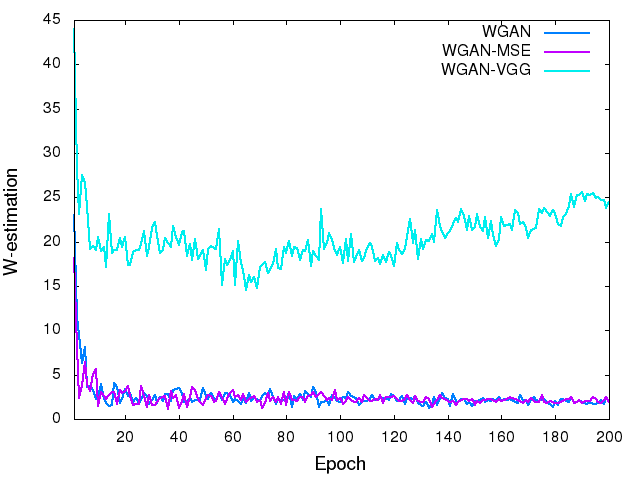}
\label{fig: w_distance_convergence}}

\caption{Plots of validation loss versus the number of epochs during the training of the 5 networks. (a) MSE loss convergence, (b) VGG loss convergence and (c) Wasserstein estimation convergence.} 
\label{fig: validation_loss}

\end{figure}

\section{Experiments}
\label{sec:exp}

\subsection{Experimental Datasets}
We used a real clinical dataset authorized for \textit{``the 2016 NIH-AAPM-Mayo Clinic Low Dose CT Grand Challenge"} by Mayo Clinic for the training and evaluation of the proposed networks \cite{lowdosectgrandchallenge}. The dataset contains 10 anonymous patients' normal-dose abdominal CT images and simulated quarter-dose CT images. In our experiments, we randomly extracted 100,096 pairs of image patches from 4,000 CT images as our training inputs and labels. The patch size is $64\times64$. Also, we extracted 5,056 pairs of patches from another 2,000 images for validation. When choosing the image patches, we excluded image patches that were mostly air. For comparison, we implemented a state-of-the-art 3D dictionary learning reconstruction technique as a representative IR algorithm \cite{xu2012low,zhang2017tensor}. The dictionary learning reconstruction was performed from the LDCT projection data provided by Mayo Clinic.

\subsection{Network Training}
In our experiments, all the networks were optimized using Adam algorithm \cite{kingma2014adam}. The optimization procedure for WGAN-VGG network is shown in Fig.~\ref{Alg1}. The mini-batch size was 128. The hyper-parameters for Adam were set as $\alpha=1e-5, \beta_1=0.5, \beta_2=0.9$, and we chose $\lambda=10$ as suggested in~\cite{gulrajani2017improved}, $\lambda_1=0.1, \lambda_2=0.1$  according to our experimental experience. The optimization processes for WGAN-MSE and WGAN are similar except that line 12 was changed to the corresponding loss function, and for CNN-MSE and CNN-VGG, lines 2-10 were removed and line 12 was changed according to their loss functions.

The networks were implemented in Python with the Tensorflow library\cite{abadi2016tensorflow}. A NVIDIA Titan XP GPU was used in this study.

\subsection{Network Convergence}
To visualize the convergence of the networks, we calculated the MSE loss and VGG loss over the 5,056 image patches for validation according to Eqs.~(\ref{eq: mse_loss}) and~(\ref{eq: vgg_loss}) after each epoch.
Fig.~\ref{fig: validation_loss} shows the averaged MSE and VGG losses respectively versus the number of epochs for the five networks. Even though these two loss functions were not used at the same time for a given network, we still want to see how their values change during the training. In the two figures, both the MSE and VGG losses decreased initially, which indicates that the two metrics are positively correlated. However, the loss values of the networks in terms of MSE are increasing in the following order, CNN-MSE$<$WGAN-MSE$<$WGAN-VGG$<$CNN-VGG (Fig.~\ref{fig: mse_loss_convergence}), yet the VGG loss are in the opposite order (Fig.~\ref{fig: vgg_loss_convergence}). The MSE and VGG losses of GAN network are oscillating in the converging process. WGAN-VGG and CNN-VGG have very close VGG loss values, while their MSE losses are quite different. On the other hand, WGAN perturbed the convergence as measured by MSE but smoothly converged in terms of VGG loss. These observations suggest that the two metrics have different focuses when being used by the networks. The difference between MSE and VGG losses will be further revealed in the output images of the generators.

In order to show the convergence of WGAN part, we plotted the estimated Wasserstein values defined as $|-\mathbb{E}[D(\bm{x})]+\mathbb{E}[D(G(\bm{z}))]|$ in Eq.~(\ref{eq: loss_DG_WGAN}).  It can be observed in Fig. 4(c) that increasing the number of epochs did reduce the W-distance, although the decay rate becomes smaller. For the WGAN-VGG curve, the introduction of VGG loss has helped to improve the perception/visibility at a cost of a compromised loss measure. For the WGAN and WGAN-MSE curves, we would like to note that what we computed is a surrogate for the W-distance which has not been normalized by the total number of pixels, and if we had done such a normalization the curves would have gone down closely to zero after 100 epochs.

\begin{figure}[!t]
\centering
\subfloat[Full Dose FBP]{\includegraphics[width=1.15in]{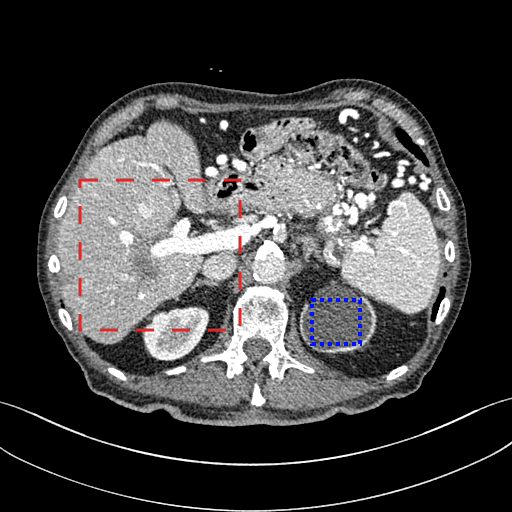}}
\subfloat[Quarter Dose FBP]{\includegraphics[width=1.15in]{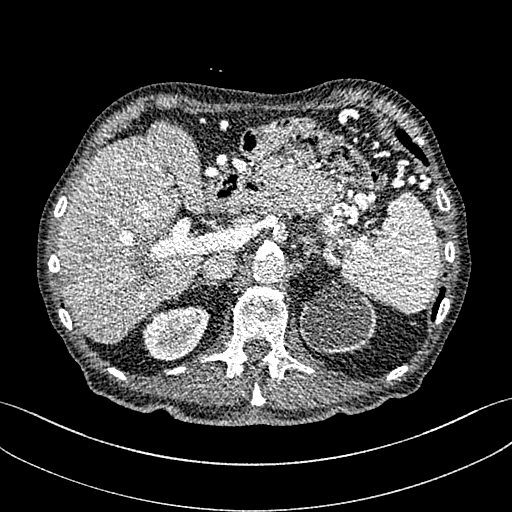}}
\subfloat[DictRecon]{\includegraphics[width=1.15in]{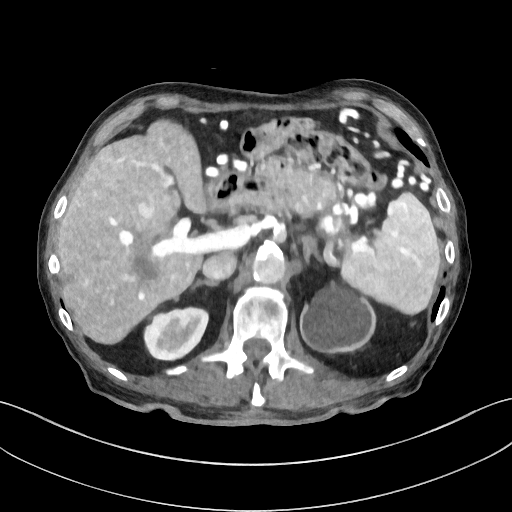}}

\subfloat[GAN]{\includegraphics[width=1.15in]{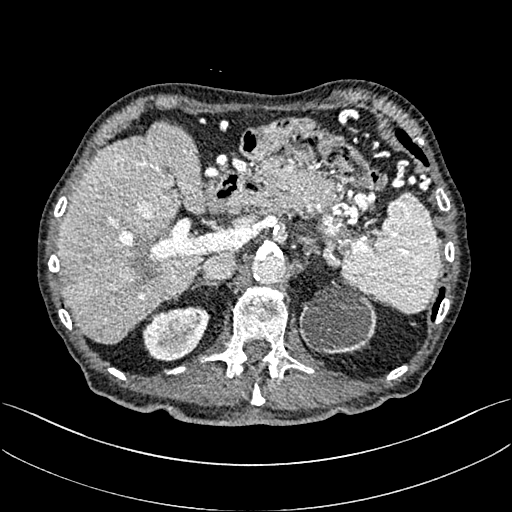}}
\subfloat[CNN-MSE]{\includegraphics[width=1.15in]{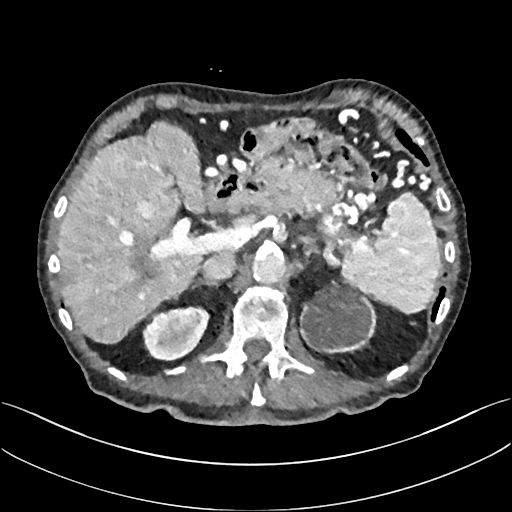}}
\subfloat[CNN-VGG]{\includegraphics[width=1.15in]{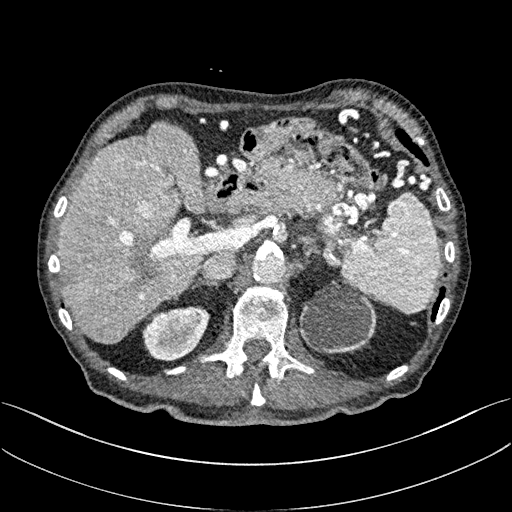}}

\subfloat[WGAN]{\includegraphics[width=1.15in]{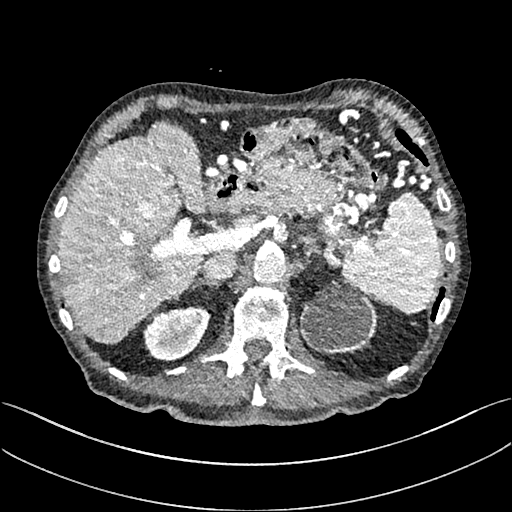}}
\subfloat[WGAN-MSE]{\includegraphics[width=1.15in]{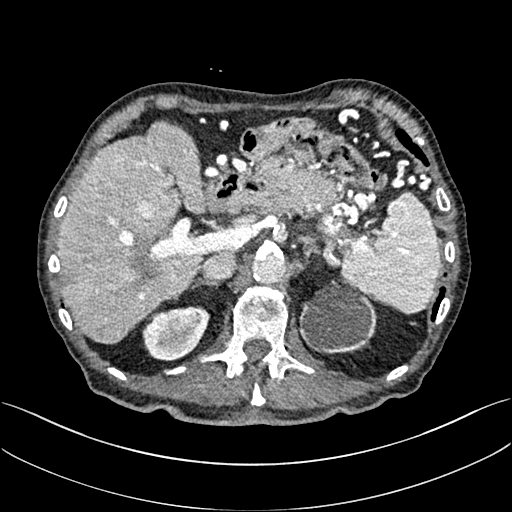}}
\subfloat[WGAN-VGG]{\includegraphics[width=1.15in]{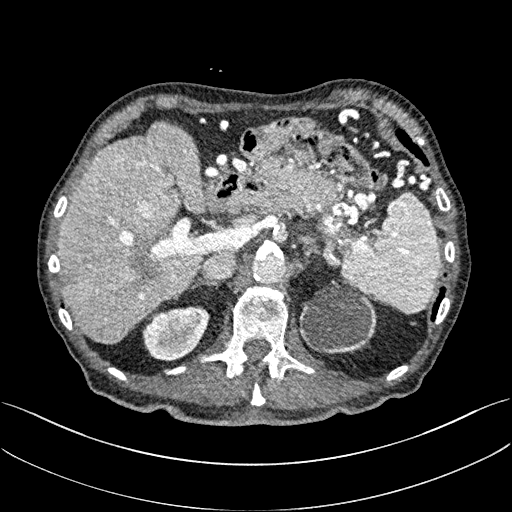}}

\caption{Transverse CT images of the abdomen demonstrate a low attenuation liver lesion (in the red box) and a cystic lesion in the upper pole of the left kidney (in the blue box). 
This display window is [-160, 240]HU.}
\label{fig: example1}
\end{figure}

\begin{figure}[!t]
\centering
\subfloat[Full Dose FBP]{\includegraphics[width=1.15in]{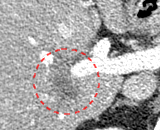}}
\subfloat[Quarter Dose FBP]{\includegraphics[width=1.15in]{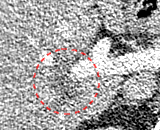}}
\subfloat[DictRecon]{\includegraphics[width=1.15in]{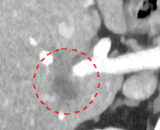}}

\subfloat[GAN]{\includegraphics[width=1.15in]{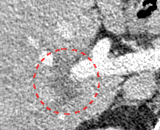}}
\subfloat[CNN-MSE]{\includegraphics[width=1.15in]{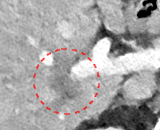}\label{fig: cnn_mse_roi1}}
\subfloat[CNN-VGG]{\includegraphics[width=1.15in]{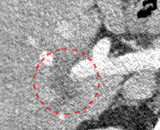}}

\subfloat[WGAN]{\includegraphics[width=1.15in]{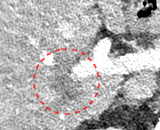}\label{fig:wgan_gp_roi1}}
\subfloat[WGAN-MSE]{\includegraphics[width=1.15in]{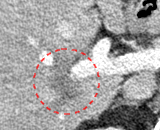}}
\subfloat[WGAN-VGG]{\includegraphics[width=1.15in]{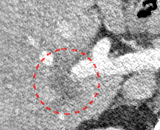}}

\caption{Zoomed ROI of the red rectangle in Fig.~\ref{fig: example1}. The low attenuation liver lesion with in the dashed circle represents metastasis. The lesion is difficult to assess on quarter dose FBP recon (b) due to high noise content. 
This display window is [-160, 240]HU.}
\label{fig: roi1}
\end{figure}

\begin{figure}[!t]
\centering
\subfloat[Full Dose FBP]{\includegraphics[width=1.15in]{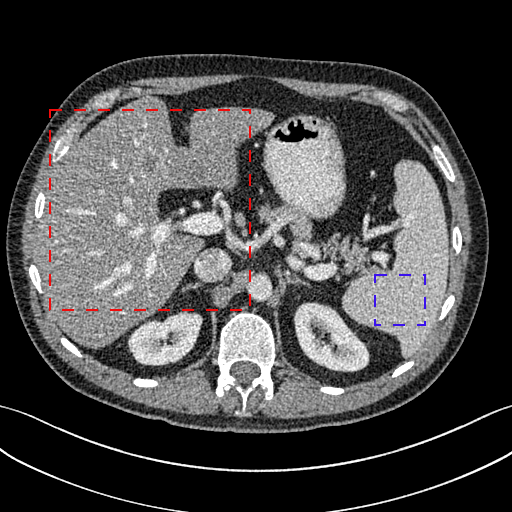}}
\subfloat[Quarter Dose FBP]{\includegraphics[width=1.15in]{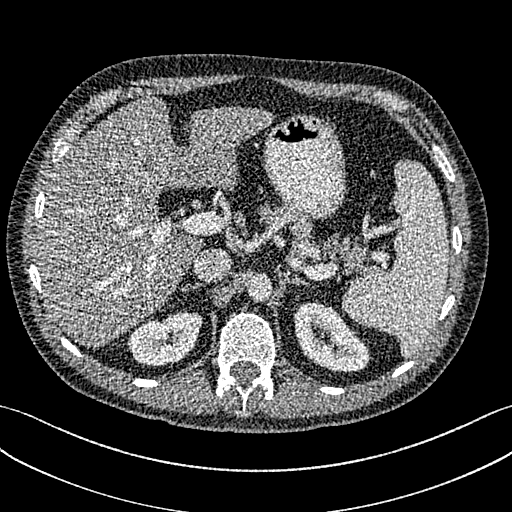}}
\subfloat[DictRecon]{\includegraphics[width=1.15in]{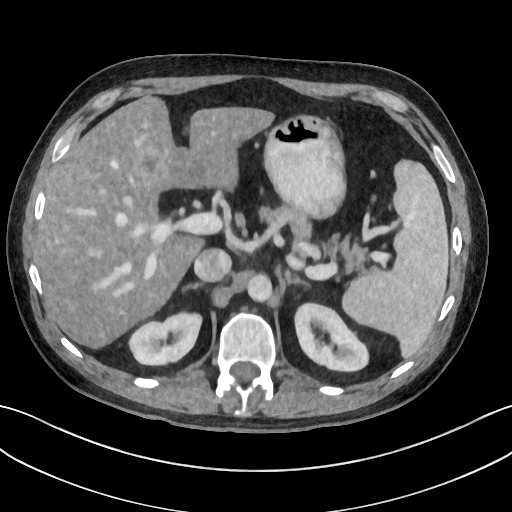}}

\subfloat[GAN]{\includegraphics[width=1.15in]{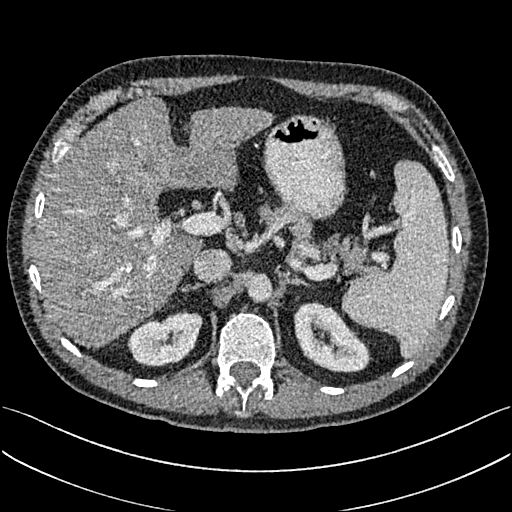}}
\subfloat[CNN-MSE]{\includegraphics[width=1.15in]{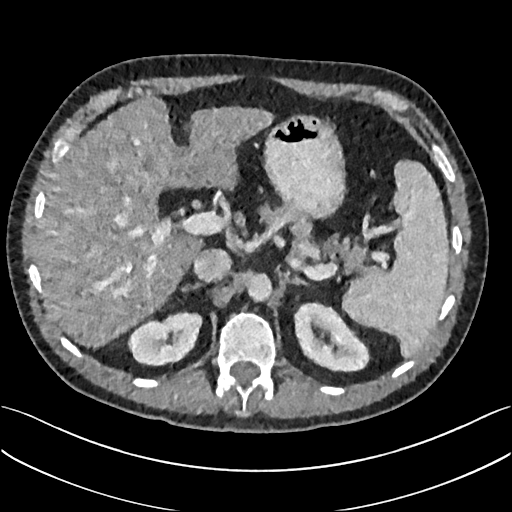}}
\subfloat[CNN-VGG]{\includegraphics[width=1.15in]{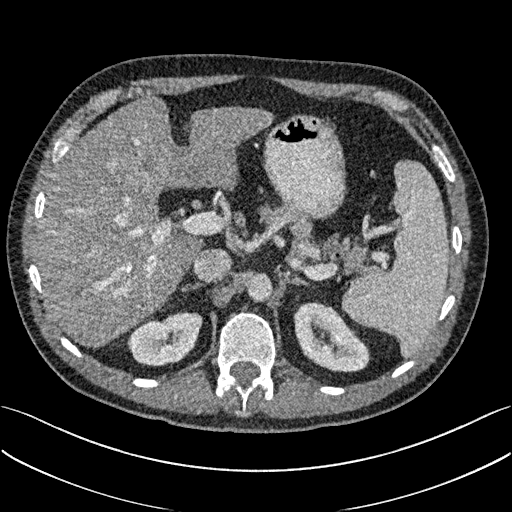}}

\subfloat[WGAN]{\includegraphics[width=1.15in]{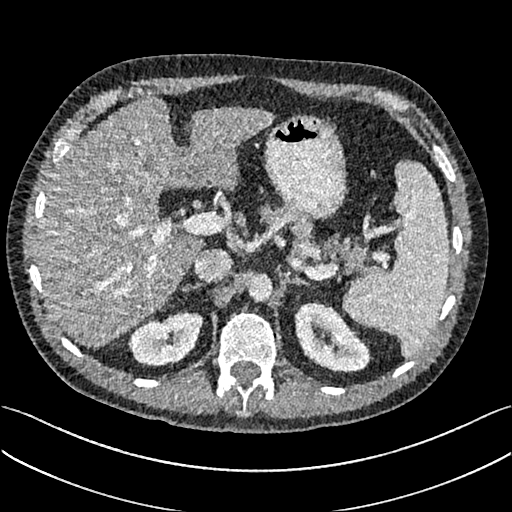}}
\subfloat[WGAN-MSE]{\includegraphics[width=1.15in]{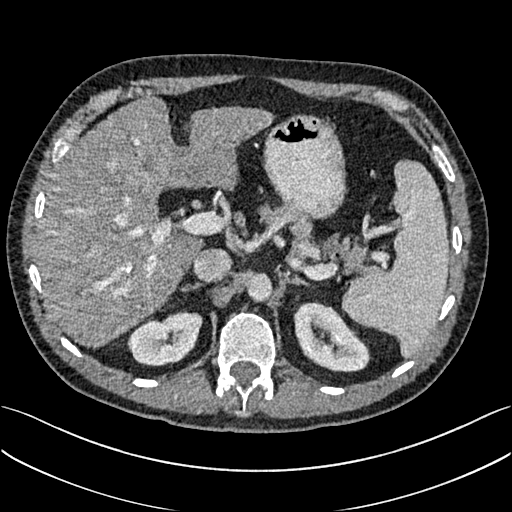}}
\subfloat[WGAN-VGG]{\includegraphics[width=1.15in]{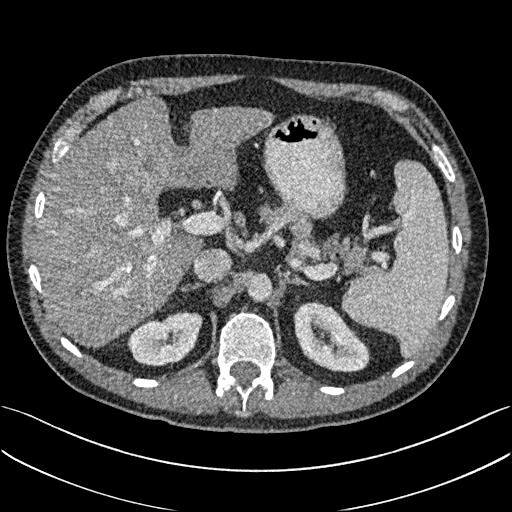}}
\caption{Transverse CT images of the abdomen demonstrate small low attenuation liver lesions. 
The display window is [-160, 240]HU.}
\label{fig: example2}
\end{figure}

\begin{figure}[!t]
\centering
\subfloat[Full Dose FBP]{\includegraphics[width=1.15in]{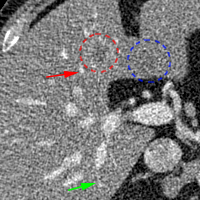}}
\subfloat[Quarter Dose FBP]{\includegraphics[width=1.15in]{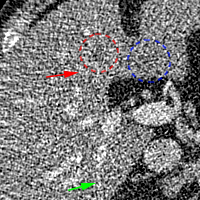}}
\subfloat[DictRecon]{\includegraphics[width=1.15in]{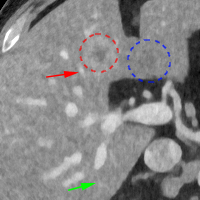}}

\subfloat[GAN]{\includegraphics[width=1.15in]{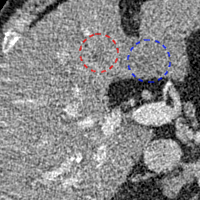}}
\subfloat[CNN-MSE]{\includegraphics[width=1.15in]{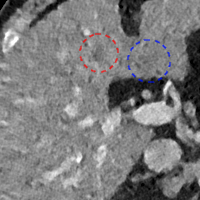}\label{fig: cnn_mse_roi2}}
\subfloat[CNN-VGG]{\includegraphics[width=1.15in]{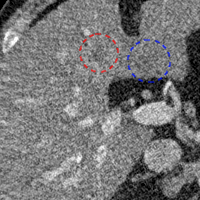}}

\subfloat[WGAN]{\includegraphics[width=1.15in]{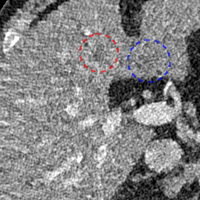}\label{fig:wgan_gp_roi2}}
\subfloat[WGAN-MSE]{\includegraphics[width=1.15in]{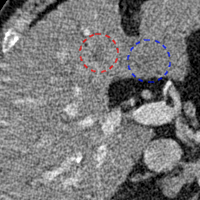}}
\subfloat[WGAN-VGG]{\includegraphics[width=1.15in]{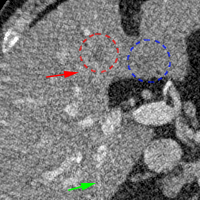}}

\caption{Zoomed ROI of the red rectangle in Fig.~\ref{fig: example2} demonstrates the two attenuation liver lesions in the red and blue circles. 
The display window is [-160, 240]HU. }
\label{fig: roi2}
\end{figure}

\subsection{Denoising Results}

To show the denoising effect of the selected networks, we took two representative slices as shown in Figs.~\ref{fig: example1} and~\ref{fig: example2}. And Figs.~\ref{fig: roi1} and~\ref{fig: roi2} are the zoomed regions-of-interest (ROIs) marked by the red rectangles in Figs.~\ref{fig: example1} and~\ref{fig: example2}. All the networks demonstrated certain denoising capabilities. However, CNN-MSE blurred the images and introduced waxy artifacts as expected, which are easily observed in the zoomed ROIs in Figs.~\ref{fig: cnn_mse_roi1} and~\ref{fig: cnn_mse_roi2}. WGAN-MSE was able to improve the result of CNN-MSE by  avoiding over-smooth but minor streak artifacts can still be observed especially compared to CNN-VGG and WGAN-VGG. Meanwhile, using WGAN or GAN alone generated stronger noise (Figs.~\ref{fig:wgan_gp_roi1} and~\ref{fig:wgan_gp_roi2}) than the other networks enhanced a few white structures in the WGAN/GAN generated images, which are originated from the low dose streak artifact in LDCT images, while on the contrary the CNN-VGG and WGAN-VGG images are visually more similar to the NDCT images. This is because the VGG loss used in CNN-VGG and WGAN-VGG is computed in a feature space that is trained previously on a very large natural image dataset \cite{imagenet_cvpr09}. By using VGG loss, we transferred the knowledge of human perception that is embedded in VGG network to CT image quality evaluation. The performance of using WGAN or GAN alone is not acceptable because it only maps the data distribution from LDCT to NDCT but does not guarantee the image content correspondence. As for the lesion detection in these two slices, all the networks enhance the lesion visibility compared to the original noisy low dose FBP images as noise is reduced by the different approaches.

As for iterative reconstruction technique, the reconstruction results depend greatly on the choices of the regularization parameters. The implemented dictionary learning reconstruction (DictRecon) result gave the most aggressive noise reduction effect compared to the network outputs as a result of strong regularization. However, it over-smoothed some fine structures. For example, in Fig.~\ref{fig: roi2}, the vessel pointed by the green arrow was smeared out while it is easily identifiable in NDCT as well as WGAN-VGG images. Yet, as an iterative reconstruction method, DictRecon has its advantage over post-processing method. As pointed by the red arrow in Fig~\ref{fig: roi2}, there is a bright spot which can be seen in DictRecon and NDCT images, but is not observable in LDCT and network processed images. Since the WGAN-VGG image is generated from LDCT image, in which this bright spot is not easily observed, it is reasonable that we do not see the bright spot in the images processed by neural networks. In other words, we do not want the network to generate structure that does not exist in the original images. In short, the proposed WGAN-VGG network is a post-processing method and information that is lost during the FBP reconstruction cannot easily be recovered, which is one limitation for all the post-processing methods. On the other hand, as an iterative reconstruction method, DictRecon algorithm generates images from raw data, which has more information than the post-processing methods.

\begin{table}[t]
\renewcommand{\arraystretch}{1.3}
\centering
\caption{Quantitative results associated with different network outputs for Figs.~\ref{fig: example1} and~\ref{fig: example2}}
\begin{tabular}{c c c c c c}
\hline
& \multicolumn{2}{c}{Fig.~\ref{fig: example1}} && \multicolumn{2}{c}{Fig.~\ref{fig: example2}} \\
& PSNR & SSIM && PSNR & SSIM \\
\cline{2-3}\cline{5-6}
LDCT & 19.7904 & 0.7496 && 18.4519 & 0.6471 \\
CNN-MSE & 24.4894 & 0.7966 && 23.2649 & 0.7022 \\
WGAN-MSE & 24.0637 & 0.8090 && 22.7255 & 0.7122 \\
CNN-VGG & 23.2322 & 0.7926 && 22.0950 & 0.6972 \\
WGAN-VGG & 23.3942 & 0.7923 && 22.1620 & 0.6949 \\
WGAN & 22.0168 & 0.7745 && 20.9051 & 0.6759 \\`1	
GAN & 21.8676 & 0.7581 && 21.0042 & 0.6632 \\
DictRecon & 24.2516 & 0.8148 && 24.0992 & 0.7631 \\
\hline
\end{tabular}
\label{table: psnr&ssim}
\end{table}

\begin{table}[!t]
\renewcommand{\arraystretch}{1.3}
\centering
\caption{statistical properties of the blue rectangle areas in  Figs.~\ref{fig: example1} and~\ref{fig: example2}. The values are in Hounsfield Unit (HU).}
\begin{tabular}{c c c c c c}
\hline
& \multicolumn{2}{c}{Fig.~\ref{fig: example1}} && \multicolumn{2}{c}{Fig.~\ref{fig: example2}} \\
& Mean & SD && Mean & SD \\
\cline{2-3}\cline{5-6}
NDCT & 9 & 36 && 118 & 38 \\
LDCT & 11 & 74 && 118 & 66 \\
CNN-MSE & 12 & 18 && 120 & 15 \\
WGAN-MSE & 9 & 28 && 115 & 25 \\
CNN-VGG & 4 & 30 && 104 & 28 \\
WGAN-VGG & 9 & 31 && 111 & 29 \\
WGAN & 23 & 37 && 135 & 33 \\
GAN & 8 & 35 && 110 & 32 \\
DictRecon & 4 & 11 && 111 & 13 \\
\hline
\end{tabular}
\label{table: stats}
\end{table}

\begin{table*}
\centering
\caption{subjective quality scores (mean$\pm$sd) for different algorithms}
\begin{adjustbox}{max width=\textwidth}
\begin{tabular}{l c c c c c c c c c}
\hline
& NDCT & LDCT & CNN-MSE & CNN-VGG & WGAN-MSE & WGAN-VGG & WGAN & GAN & DictRecon \\
\hline
Noise Suppression &  -	& -	& $4.35\pm0.24$ & $3.10\pm0.23$ & $3.55\pm0.25$	& $3.20\pm0.25$ & $2.90\pm0.26$ & $3.00\pm0.21$	& $\bm{4.65\pm0.20}$ \\
Artifact Reduction & -	& - &	$1.70\pm0.28$	& $2.85\pm0.32$ &	$3.05\pm0.27$ & 	$\bm{3.45\pm0.25}$ &	$2.90\pm0.28$ &	$3.05\pm0.27$ &	$2.05\pm0.27$ \\
Overall Quality & $3.95\pm0.20$ &	$1.35\pm0.16$	& $2.15\pm0.25$	& $3.05\pm0.20$	& $3.30\pm0.21$	& $\bm{3.70\pm0.15}$	& $3.05\pm0.22$	& $3.10\pm0.21$	& $2.05\pm0.36$ \\
\hline
\end{tabular}
\end{adjustbox}
\label{table: reader_study}
\end{table*}
\subsection{Quantitative Analysis}
For quantitative analysis, we calculated the peak-to-noise ratio (PSNR) and structural similarity (SSIM). The summary data are in Table~\ref{table: psnr&ssim}. CNN-MSE ranks the first in terms of PSNR, while WGAN is the worst. Since PSNR is equivalent to the per-pixel loss, it is not surprising that CNN-MSE, which was trained to minimize MSE loss, outperformed the networks trained to minimize other feature-based loss. It is worth noting that these quantitative results are in decent agreement with Fig.~\ref{fig: validation_loss}, in which CNN-MSE has the smallest MSE loss and WGAN has the largest. The reason why WGAN ranks the worst in PSNR and SSIM is because it does not include either MSE or VGG regularization. DictRecon achieves the best SSIM and a high PSNR. However, it has the problem of image blurring and leads to blocky and waxy artifacts in the resultant images. This indicates that PSNR and SSIM may not be sufficient in evaluating image quality.

In the reviewing process, we found two papers using similar network structures. In~\cite{wolterink2017generative}, Wolterink \textit{et al.} trained three networks, i.e. GAN, CNN-MSE, and GAN-MSE for cardiac CT denoising. Their quantitative PSNR results are consistent with our counterpart results. And Yu \textit{et al.}~\cite{yu2017deep} used GAN-VGG to handle the de-alising problem for fast CS-MRI. Their results are also consistent with ours. Interestingly, despite the high PSNRs obtained by MSE-based networks, the authors in the two papers all claim that GAN and VGG loss based networks have better image quality and diagnostic information.

To gain more insight into the output images from different approaches, we inspect the statistical properties by calculating the mean CT numbers (Hounsfield Units) and standard deviations (SDs) of two flat regions in Figs.~\ref{fig: example1} and~\ref{fig: example2} (marked by the blue rectangles). In an ideal scenario, a noise reduction algorithm should achieve mean and SD to the gold standard as close as possible. In our experiments, the NDCT FBP images were used as gold standard because they have the best image quality in this dataset. As shown in Table~\ref{table: stats}, Both CNN-MSE and DictRecon produced much smaller SDs compared to NDCT, which indicates they over-smoothed the images and supports our visual observation. On the contrary, WGAN produced the closest SDs yet smaller mean values, which means it can reduce noise to the same level as NDCT but it compromised the information content. On the other hand, the proposed WGAN-VGG has outperformed CNN-VGG, WGAN-MSE and other selected methods in terms of mean CT numbers, SDs, and most importantly visual impression.

In addition, we performed a blind reader study on 10 groups of images. Each group contains the same image slice but processed by different methods. NDCT and LDCT images are also included for reference, which are the only two labeled images in each group. Two radiologists were asked to independently score each image in terms of noise suppression and artifact reduction on a five-point scale (1 = unacceptable and 5 = excellent), except for the NDCT and LDCT images, which are the references. In addition, they were asked to give an overall image quality score for all the images. The mean and standard deviation values of the scores from the two radiologists were then obtained as the final evaluation results, which are shown in Table.~\ref{table: reader_study}. It can be seen that CNN-MSE and DictRecon give the best noise suppression scores while the proposed WGAN-VGG outperforms the other methods for artifact reduction and overall quality improvement. Also, *-VGG networks provide higher scores than *-MSE networks in terms of artifact reduction and overall quality but lower scores for noise suppression. This indicates that MSE loss based networks are good at noise suppression at a loss of image details, resulting in an image quality degradation for diagnosis. Meanwhile, the networks using WGAN give better overall image quality than the networks using CNN, which supports the use of WGAN for CT image denoising.

\begin{figure}[!t]
\centering
\subfloat[VGG Map of Full Dose Image]{\includegraphics[width=3.4in]{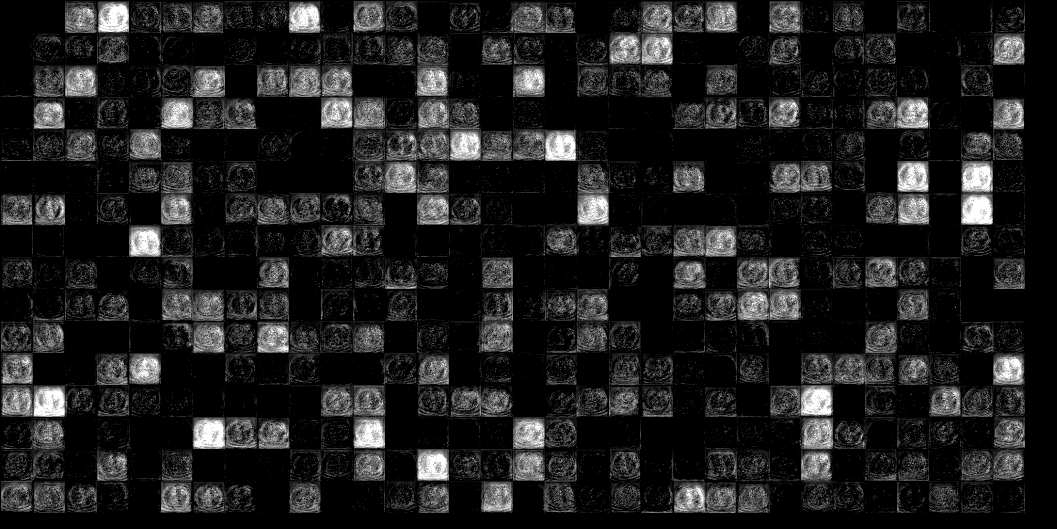}}

\subfloat[VGG Map of Quarter Dose Image]{\includegraphics[width=3.4in]{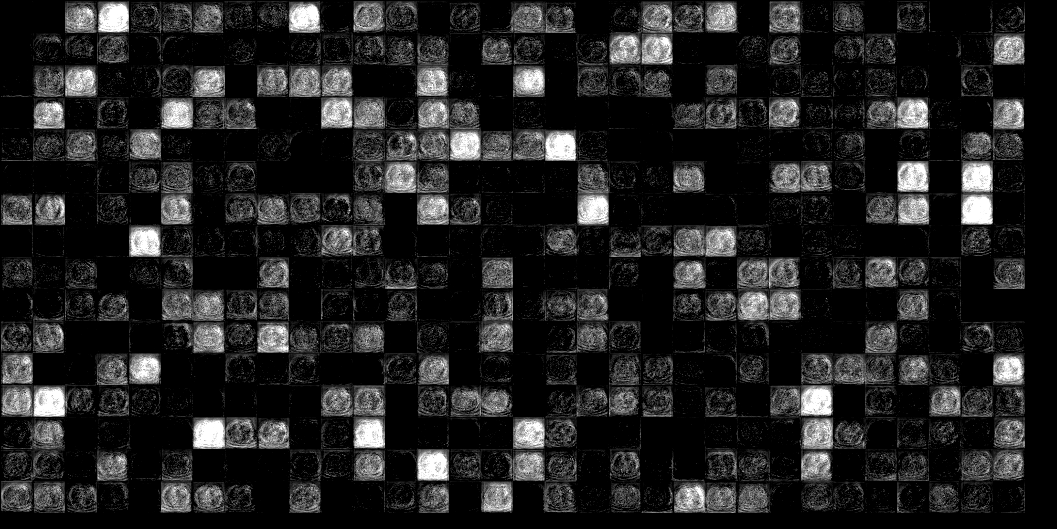}}

\subfloat[Absolute Difference]{\includegraphics[width=3.4in]{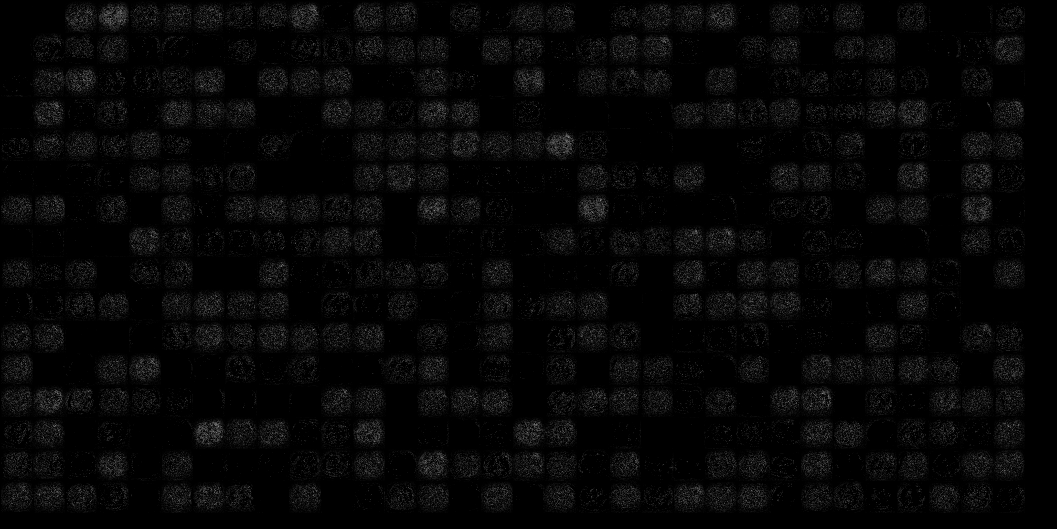}}

\caption{VGG feature maps of full dose and quarter dose images in Fig.~\ref{fig: example1} and their absolute difference.} 
\label{fig: vgg_maps}
\end{figure}
\subsection{VGG Feature Extractor}

Since VGG network is trained on natural images, it may cause concerns on how well it performs on CT image feature extraction. Thus, we displayed two feature maps of normal dose and quarter dose images and their absolute difference in Fig.~\ref{fig: vgg_maps}. The feature map contains 512 small images of size $32\times32$. We organize these small images into a $32\times16$ array. Each small image emphasizes a feature of the original CT image, i.e. boundaries, edges, or whole structures. Thus, we believe VGG network can also serve a good feature extractor for CT images.

\section{Discussions and Conclusion}
\label{sec:conclusions}

The most important motivation for this paper is to approach the gold standard NDCT images as much as possible. As described above, the feasibility and merits of GAN has been investigated for this purpose with the Wasserstein distance and the VGG loss. The difference between using the MSE and VGG losses is rather significant. Despite the fact that networks with MSE would offer higher values for traditional figures of merit, VGG loss based networks seem desirable for better visual image quality with more details and less artifacts.

The experimental results have demonstrated that using WGAN helps improve image quality and statistical properties. Comparing the images of CNN-MSE and WGAN-MSE, we can see that the WGAN framework helped to avoid over-smoothing effect typically suffered by MSE based image generators. Although CNN-VGG and WGAN-VGG visually share a similar result, the quantitative analysis shows WGAN-VGG enjoys higher PSNRs and more faithful statistical properties of denoised images relative to those of NDCT images. However, using WGAN/GAN alone reduced noise but at the expense of losing critical features. The resultant images do not show a strong noise reduction. Quantitatively, the associated PSNR and SSIM increased modestly compared to LDCT but they are much lower than what the other networks produced. Theoretically, WGAN/GAN network is based on generative model and may generate images that look naturally yet cause a severe distortion for medical diagnostics. This is why an additive loss function such as MSE and VGG loss should be added to guarantee the image content remains the same.

It should be noted that the experimental data contain only one noise setting. Networks should be re-trained or re-tuned for different data to adapt for different noise properties. Especially, networks with WGAN are trying to minimize the distance between two probability distributions. Thus, their trained parameters have to be adjusted for new datasets. Meanwhile, since the loss function of WGAN-VGG is a mixture of feature domain distance and the GAN adversarial loss, they should be carefully balanced for different dataset to reduce the amount of image content alternation.

The denoising network is a typical end-to-end operation, in which the input is a LDCT image while the target is a NDCT image. Although we have generated images visually similar to NDCT counterparts in the WGAN-VGG network, we recognize that these generated images are still not as good as NDCT images. Moreover, noise still exists in NDCT images. Thus, it is possible that VGG network has captured these noise features and kept them in the denoised images. This could be a common problem for all the denoising networks. How to outperform the so-called gold standard NDCT images is an interesting open question. Moreover, image post-denoising methods also suffer from the information loss during the FBP reconstruction process. This phenomena is observed in the comparison with DictRecon result. A better way to incorporate the strong fitting capability of neural network and the data completeness of CT data is to design a network that maps directly from raw projection to the final CT images, which could be a next step of our work.

In conclusion, we have proposed a contemporary deep neural network that uses a WGAN framework with perceptual loss function for LDCT image denoising. Instead of focusing on the design of a complex network structure, we have dedicated our effort to combine synergistic loss functions that guide the denoising process so that the resultant denoised results are as close to the gold standard as possible. Our experiment results with real clinical images have shown that the proposed WGAN-VGG network can effectively solve the well-known over-smoothing problem and generate images with reduced noise and increased contrast for improved lesion detection. In the future, we plan to incorporate the WGAN-VGG network with more complicated generators such as the networks reported in~\cite{chen_zhang_kalra_lin_chen_liao_zhou_wang_2017,kang2016deep} and extend these networks for image reconstruction from raw data by making a neural network counterpart of the FBP process.

%

\bibliographystyle{IEEEtran}
\bibliography{denoisingCT}
\ifCLASSOPTIONcaptionsoff
  \newpage
\fi

\end{document}